\documentclass[10pt,twocolumn,letterpaper]{article}

\usepackage{iccv}
\usepackage{times}
\usepackage{epsfig}
\usepackage{graphicx}
\usepackage{amsmath}
\usepackage{amssymb}

\usepackage{amsmath}
\usepackage{amsfonts}
\usepackage{amssymb}
\usepackage{wrapfig}
\usepackage{subcaption}
\usepackage{multirow}
 \usepackage{mathtools} 

\usepackage{verbatim}

\usepackage{anyfontsize}
\usepackage{tcolorbox}
\usepackage{booktabs}
\usepackage{color}
\usepackage{tikz}
\usetikzlibrary{arrows,shapes,snakes,automata,backgrounds,fit,petri}
\usepackage{adjustbox}

\newcommand{\RN}[1]{%
	\textup{\lowercase\expandafter{\it \romannumeral#1}}%
}

\makeatletter
\newcommand{\distas}[1]{\mathbin{\overset{#1}{\kern\z@\sim}}}%

\usepackage{enumitem}

\usepackage[lined,boxed,commentsnumbered,ruled,linesnumbered]{algorithm2e}

\newcommand{\beq}{\vspace{0mm}\begin{equation}}
\newcommand{\eeq}{\vspace{0mm}\end{equation}}
\newcommand{\beqs}{\vspace{0mm}\begin{eqnarray}}
\newcommand{\eeqs}{\vspace{0mm}\end{eqnarray}}
\newcommand{\barr}{\begin{array}}
\newcommand{\earr}{\end{array}}

\newcommand{\E}{\mathbb{E}}

\newcommand{\Dcal}{\mathcal{D}}

\newcommand{\Tcal}{\mathcal{T}}

\newcommand{\Hcal}{\mathcal{H}}

\newcommand{\Ucal}{\mathcal{U}}

\ifx\assumption\undefined
\newtheorem{assumption}{Assumption}
\fi

\ifx\definition\undefined
\newtheorem{definition}{Definition}
\fi

\ifx\remark\undefined
\newtheorem{remark}{Remark}
\fi

\usepackage[pagebackref=true,breaklinks=true,colorlinks,bookmarks=false]{hyperref}

\iccvfinalcopy % *** Uncomment this line for the final submission

\def\iccvPaperID{****} % *** Enter the ICCV Paper ID here

\ificcvfinal\fi

\begin{document}
%%%%%%%%% TITLE
% \title{Neural Architecture Search for Out-of-Distribution Detection}
\title{Can Dense Connectivity Benefit Outlier Detection? An Odyssey with NAS}

\author{Hao Fu,  Tunhou Zhang,  Hai Li,  Yiran Chen\\
% Duke University \\
Duke University, Durham, NC 27708\\
{\tt\small \{hao.fu,tunhou.zhang,hai.li,yiran.chen\}@duke.edu}
% \and
% Tunhou Zhang\\
% Duke University\\
% % First line of institution2 address\\
% {\tt\small tunhou.zhang@duke.edu}
}

\maketitle
\ificcvfinal\thispagestyle{empty}\fi

\begin{abstract}
   % The reliability of machine learning systems depends not only on their accuracy but also on their ability to handle distribution shift and outliers.
   Recent advances in Out-of-Distribution (OOD) Detection is the driving force behind safe and reliable deployment of Convolutional Neural Networks (CNNs) in real-world applications.
   % This makes Out-of-Distribution (OOD) detection a crucial aspect of safe deployment in real-world applications. 
   However, existing studies focus on OOD detection through confidence score and deep generative model-based methods, without considering the impact of DNN structures, especially dense connectivity in architecture fabrications. 
   In addition, %confidence score detection 
   % that do not modify the training procedure have been shown to
   existing outlier detection approaches exhibit high variance in generalization performance, lacking stability and confidence in evaluating and ranking different outlier detectors.
   In this work, we propose a novel paradigm, Dense Connectivity Search of Outlier Detector (DCSOD), that automatically explore the dense connectivity of CNN architectures on near-OOD detection task using Neural Architecture Search (NAS).
   %predictor-based neural architecture search (NAS) to automatically search for optimal structure designs for near-OOD detection task. 
   % To enable comprehensive exploration of dense connection patterns
   We introduce a hierarchical search space containing versatile convolution operators and dense connectivity, allowing a flexible exploration of CNN architectures with diverse connectivity patterns. 
   To improve the quality of evaluation on OOD detection during search, we propose evolving distillation based on our multi-view feature learning explanation. Evolving distillation stabilizes training for OOD detection evaluation, thus improves the quality of search. 
   % We also identify the issue of high evaluation variance arising from training and 
   % This algorithm effectively 
   We thoroughly examine DCSOD on CIFAR benchmarks under OOD detection protocol.% CIFAR-10 vs CIFAR-100 and CIFAR-100 vs CIFAR-10 benchmarks.
   Experimental results show that DCSOD achieve remarkable performance over widely used architectures and previous NAS baselines. Notably, DCSOD achieves state-of-the-art (SOTA) performance on CIFAR benchmark, with AUROC improvement of $\sim$1.0\%.
\end{abstract}

% %-------------------------------------------------------------------------
\section{Introduction}
The success of Convolutional Neural Network (CNN) achieve record-breaking performance under various real-world applications, such as medical image analysis~\cite{weng2019unet,cao2023swin}, autonomous driving~\cite{geiger2013vision,tang2020searching}, and human pose estimation~\cite{sun2019deep,cai2020learning}.
% Modern machine learning system requires that the models not only be accurate but also robust to distribution shift and outliers. 
While modern CNN architectures enjoy high performance on computer vision benchmarks thanks to continuous development and techniques from latest research, the robustness of these models in addressing data distribution shift is rarely studied.
Thus, Out-of-Distribution (OOD) detection~\cite{yang2021generalized} becomes an essential part for safe deployment of CNN models.

Detecting OOD inputs has made significant progress in recent years. Most previous studies have focused on developing post-hoc techniques using confidence score~\cite{hendrycks2016baseline, lee2018simple, liu2020energy} or methods based on deep generative models~\cite{zhang2020hybrid}. Another branch of OOD detection methods makes use of auxiliary outlier samples with extra regularization during training~\cite{hendrycks2018deep,papadopoulos2021outlier} to help models learn ID/OOD discrepancy. However, there is an orthogonal aspect from the intrinsic influence of structure to OOD detection which still remains unexplored. The significance of network architectures in OOD detection has emerged in several previous experiments. Liang et al.~\cite{liang2017enhancing} find out different architectures with close classification accuracy on in-distribution data demonstrate clear OOD detection performance difference using MSP and ODIN method. 
Koner et al. ~\cite{koner2021oodformer} proposed a new architecture Oodformer that leverages the contextualization capabilities of the transformer to improve OOD detection performance.

% Despite the significant impact of architecture on the performance,
Recent development and evolution in CNN architectures~\cite{hu2018squeeze,tan2019mnasnet,cai2019once} significantly boost the performance on image applications with chain-structured CNN architectures. % For example, Neural Architecture Search (NAS) boosts ImageNet classification accuracy from 72\% in MobileNetV2 to 80\% in Once-for-all under mobile computation regime.
Yet, existing OOD research overlooks such advancements and instead, use conventional chain-like CNN architectures (e.g., ResNet~\cite{he2016deep}) as backbones to carry outlier detection, achieving modest AUROC gain compared to prior state-of-the-art baselines.
Thus, a systematic approach that thoroughly studies the relationship between CNN architecture fabrications and the quality outlier detectors is beneficial to push the boundary of outlier detection.

\begin{figure}
\centering
   \includegraphics[width=1\linewidth]{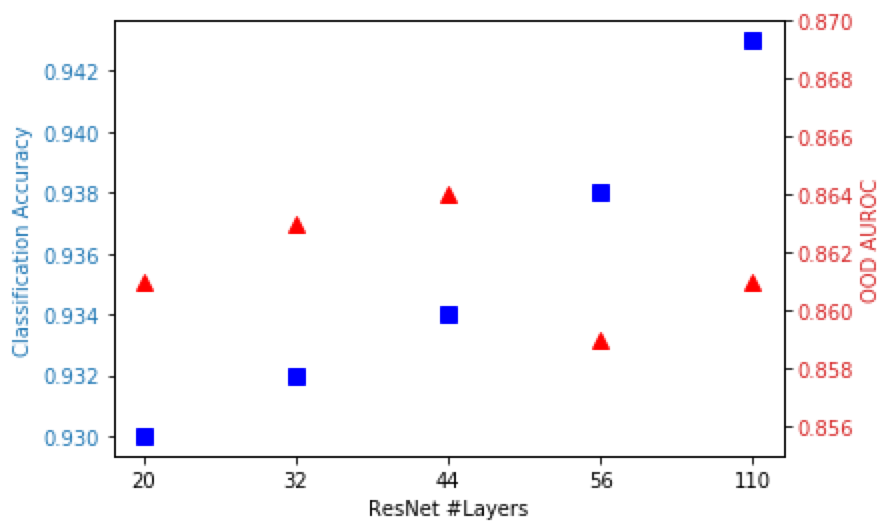}
   \caption{Classification Accuracy v.s. OOD AUROC of ResNets with different number of layers. Better accuracy does not indicate higher AUROC score.}
   \vspace{-2em}
\label{fig:resnet_depth}
\end{figure}

Motivated by the strong advancements brought by Neural Architecture Search (NAS) on image classification, a natural idea is incorporating NAS into the search of CNN architectures dedicated for outlier detection.
However, NAS has two major challenges in crafting optimal outlier detectors. First, image-based NAS approaches use hand-crafted design motifs (e.g., ResNet block~\cite{he2016deep}) as building block to craft CNN architectures, taking advantage of its strong performance on image classification. Yet, these design motifs may not prove effective on outlier detection, see Figure \ref{fig:resnet_depth}. This calls for a more flexible search space to cover more flexible CNN designs, i.e., dense connectivity of versatile convolution operators. 
Second, evaluating outlier detectors require precise techniques to distinguish the OOD performance of different NAS models. While existing OOD research made efforts to enhance the quality of outlier detector, their evaluation metric suffers from high variance and thus, not feasible to be deployed in NAS process.
This calls for a more stable and robust evaluation strategy to improve the ranking of different CNN models.
% Currently, there are no systematic methods to guide the architecture design specifically for OOD detection

Given the above challenges and motivations, our solution is establishing a systematic framework to search for the dense connectivity of versatile convolution building operators for outlier detection. 
Thus, we focus on the following question in this paper, ``\textit{Can the search of dense connectivity in CNN architectures benefit outlier detectors?}''

To fully answer the question, we propose a new paradigm, \textbf{D}ense \textbf{C}onnectivity \textbf{S}earch of \textbf{O}utlier \textbf{D}etector (DCSOD), that employ Neural Architecture Search (NAS) to automatically search for the optimal outlier detectors for out-of-distribution samples. 
We focus on the challenging near-OOD detection task, where the outliers exhibit semantic similarity to the in-distribution classes. 
% To enable comprehensive exploration of dense connection patterns within neural networks, 
We introduce a flexible convolutional neural network (CNN) search space containing dense connectivity of versatile building operators.
% We creates a hierarchical representation of the CNN backbone, consisting 
More specfically, we represent the hierarchical structure of CNN architectures as a composition of Directed Acyclic Graphs (DAGs). Each DAG represents a building cell in a particular hierarchy.
As such, our search space includes a wealth of potential candidates and encompasses a wide range of design motifs (e.g., ResNet~\cite{he2016deep}, DenseNet~\cite{huang2017densely}), allowing versatile dense connectivity between diverse computation operators with minimal structural constraints.
% The DAG representation captures the edge connections between heterogeneous building operators, including convolution and depthwise convolution with varying filter sizes. 

We obtain the optimal outlier detector from our search space by leveraging neural predictor~\cite{wen2020neural}.
We identify the challenge of high evaluation variance on OOD detection evaluation based on multi-view feature learning: The subset of features that are acquired through cross entropy training exhibit differing levels of ability in recognizing out-of-distribution (OOD) data, resulting in a significant fluctuation in the efficacy of OOD detection.
To tackle the challenge, we introduce a new outlier distillation training objective that offers more informative guidance and enables efficient multi-view learning for outlier detection, and propose an evolving distillation algorithm to mitigate the variance in OOD evaluation.
The proposed algorithm searches for an evolving ensemble of teacher models, resulting in a progressively improved training loss landscape and providing better guidance on ranking candidate CNN architectures in predictor-based NAS.

We examine the proposed methods on CIFAR benchmarks
% One of the challenges in building a neural predictor for OOD detection is the high evaluation variance that arises from training. 
% the CIFAR-10 vs CIFAR-100 and CIFAR-100 vs CIFAR-10 benchmarks 
and achieve state-of-the-art (SOTA) performance over existing CNN-based structures and NAS methods. More specifically, DCSOD achieve 95.8\% (95.1\%) AUROC score on CIFAR-10 vs. CIFAR-100 (CIFAR-100 vs. CIFAR-10) detection task, improving previous state-of-the-art by 0.8\% (1.1\%). We also conduct extensive experiments to study the effects of hyper-parameters, loss landscape structures, and extract beneficial structural patterns for outlier detection. We summarize our contributions as follows:

\begin{itemize}[noitemsep,leftmargin=*]
    \item We propose DCSOD, a novel paradigm that leverages Neural Architecture Search (NAS) to explore optimal outlier detectors within a highly flexible search space, brewing CNN architectures with dense connectivity of versatile convolution operators.
    \item We provide an interpretation based on multi-view feature learning and propose an outlier distillation loss along with an evolving distillation algorithm to stabilize training and reduce variance in OOD detection evaluations.
    \item The best models crafted by DCSOD, achieve the state-of-the-art OOD AUROC on CIFAR benchmarks.
\end{itemize}
% %-------------------------------------------------------------------------
\section{Related Work}
\noindent \textbf{OOD Detection.}
Most popular techniques for OOD detection involve designing a confidence score to distinguish outlier data. One of the most basic methods for OOD detection is to use the maximum softmax probability (MSP)\cite{hendrycks2016baseline} as the confidence score. Lee et al.\cite{lee2018simple} proposed fitting a class-conditional Gaussian distribution with feature embeddings and using the Mahalanobis distance (Maha) for OOD detection. Another line of work focus on modifying the training loss with usage of outlier examples. For instance, Outlier Exposure (OE)\cite{hendrycks2018deep} trains the model to predict a uniform distribution over in-distribution classes. Liu et al. \cite{liu2020energy} proposed energy-bounded learning objective to explicitly shape the energy surface for differentiation between in- and out-of-distribution data.

\begin{figure*}[t]
\begin{center}
    \includegraphics[width=1\linewidth]{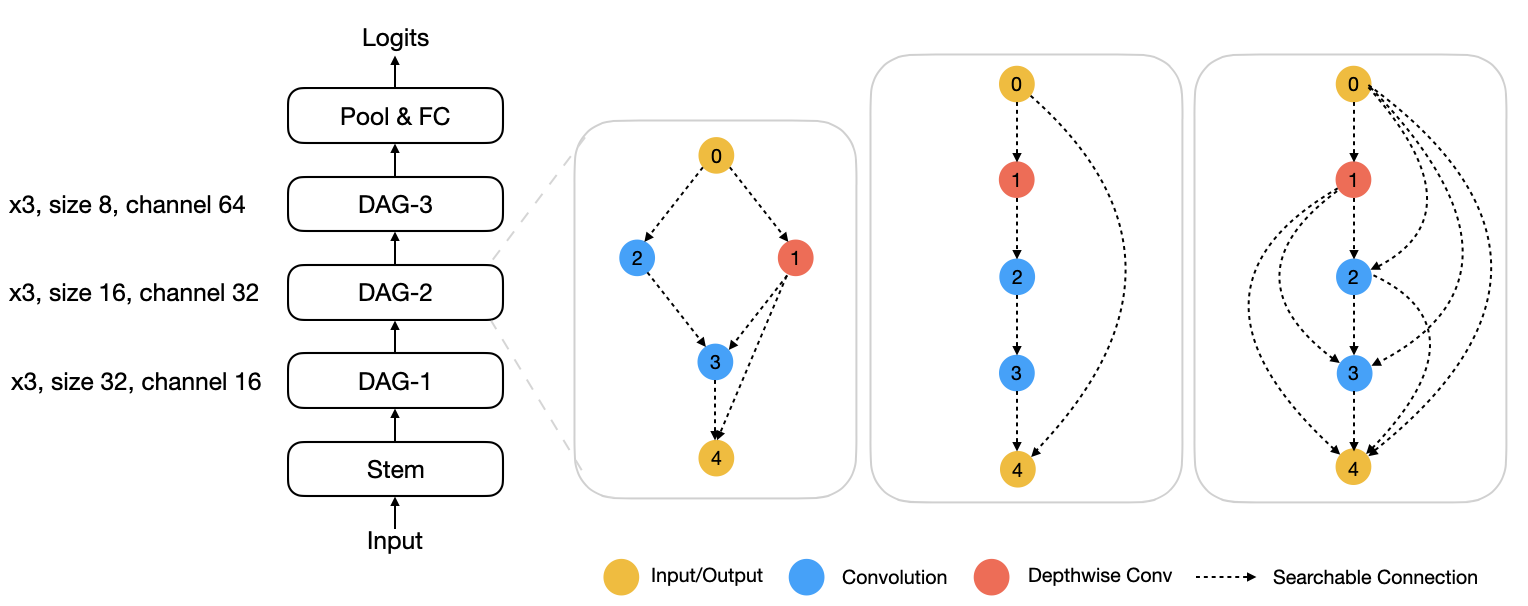}
   \caption{The dense connectivity search space, with three examples of dense connectivity in a cell. Standard structure such as ResNet and DenseNet are included.}
\label{fig:cifar_dag}
\end{center}
\vspace{-2em}

\end{figure*}

\noindent \textbf{Neural Architecture Search (NAS).}
Neural Architecture Search (NAS) democratizes the development of Convolutional Neural Network (CNN) models on computer vision applications~\cite{zoph2018learning,sun2019deep,liu2019auto}. Reinforcement Learning~\cite{zoph2018learning,pham2018efficient}, differentiable-based search~\cite{liu2018darts,wu2019fbnet}, predictor-based~\cite{wen2020neural}, and one-shot search~\cite{cai2019once,yu2020bignas} are popular search methods in delivering state-of-the-art vision models.
However, most existing NAS methods utilizes design motifs as building blocks to brew a limited search space, with strong human priors on the quality of image perception. Yet, the same prior may not be optimal on outlier detection problems dealing with data distribution.
A few NAS works attempt to explore the structural wiring of CNN architectures~\cite{xie2019exploring,luo2018neural,liu2017hierarchical}, yet end up with constraints (e.g., limitation of building operators between nodes) in search space and modest performance on vision benchmarks.
Our approach extends the search space to enable dense connectivity between different convolutions with minimal constraints in search space, thus provides more opportunity in obtaining better outlier detectors.
% a. one-shot, predictor based

%\subsection{Knowledge Distillation} 
% a. Multi-view feature learning beneficial for OOD detection
% %-------------------------------------------------------------------------
\section{Methodology}
The relationship between CNN architectures and OOD evaluation quality is left unexplored in existing literature.
In this work, we propose a novel paradigm, \textbf{D}ense \textbf{C}onnectivity \textbf{S}earch of \textbf{O}utlier \textbf{D}etector (DCSOD), to explore optimal outlier detectors using Neural Architecture Search.
DCSOD searches for the optimal structural wiring of convolution operators within a dense connectivity search space, see Figure \ref{fig:cifar_dag}.
The dense connectivity between different convolution operators allow information to be aggregated without structural constraints, thus extending the flexibility in CNN architecture designs compared to chain-structured counterparts (e.g., ResNet).
DCSOD utilizes a neural predictor~\cite{wen2020neural} to explore the design space, and introduces evolving distillation algorithm to enhance the ranking of CNN architectures. The evolving distillation algorithm enables efficient multi-view learning for outlier detection, thus obtains a more precise and robust OOD evaluation.

\subsection{NAS Framework for OOD Detection}
In this section, we introduce the dense connectivity space and predictor-based NAS.

\noindent \textbf{Dense Connectivity Search Space.}
Modern CNN architectures are composed of a series of building blocks (i.e. stages), where each stage is responsible for transforming features at the same input resolution. Our search space follows the same paradigm but focuses on structural design in terms of connectivity between operators inside each stage. As shown in Figure~\ref{fig:cifar_dag}, the dense connectivity search space has three stages, each consisting of three identical cells. The cells from different stages are independent and can be different. We represent each building cell as a Directed Acyclic Graph (DAG) to capture the edge connections between diversified vertices (i.e., operators). Each vertex in the DAG is an atomic convolutional operator selected from \{Convolution 1$\times$1, 3$\times$3, 5$\times$5\} and \{Depthwise Convolution 3$\times$3, 5$\times$5\}. This constitutes a large search space covering several standard architectures, such as ResNet, DenseNet, etc.

 Specifically, among all vertices in a DAG from stage $s$ defined as $\mathcal{G}^{(s)}=(\mathcal{V}^{(s)}, \mathcal{E}^{(s)})$, $v_0$ is the input that receives the output from the previous cell, and $v_K$ is the output that concatenate all leaf vertices in the DAG. Each intermediate vertex $v_i \in \{v_1,v_2...,v_{K-1}\}$ can connect to its preceding vertices and produce the output $X_{v_i}$ with the assigned operators $op^{(i)}$:
\begin{align}
	X_{v_i}=op^{(i)}(Concat_{\{v_j | j<i, e_{ji}\in\mathcal{E}^{(k)}\}}[X_{v_j}])
\end{align}
Given the selected $K$ vertex operators, the entire architecture $\mathcal{A}$ can then be represented as the union of independent DAGs, i.e. $\mathcal{A}=\mathrm{Arch}(\mathcal{G}^{(1)}, \mathcal{G}^{(2)}, \mathcal{G}^{(3)}; op^{(1)},...op^{(K)})$ and the searching goal is to find the optimal dense connectivity inside each DAG as follows:
\begin{align}
	\mathcal{E}^* = \arg \max_{\mathcal{E}} \mathrm{Val}(\mathrm{Arch}(\mathcal{G}^{(1)}, \mathcal{G}^{(2)}, \mathcal{G}^{(3)}; op^{(1)},...op^{(K)})).
\end{align}
 where $\mathrm{Val}(\cdot)$ is the validation function. In practice, we set $K$ exceed the number of operator choices to enable more versatile dense connection. The total search space contains $2^{\frac{3K(K-1)}{2}}$ unique architectures.

\noindent \textbf{Predictor-based NAS Framework.}
We utilize predictor-based NAS\cite{wen2020neural} to systematically explore beneficial structure patterns for OOD detection. Compared to other popular NAS methods that use weight sharing techniques, predictor-based NAS uses a neural predictor as a surrogate for ground-truth performance. This approach avoids co-adaptation issues and provides trustworthy predictive performance, demonstrated as follows:
\begin{enumerate}[noitemsep,leftmargin=*]
 \item Randomly sample $N$ architecture from the search space, train and evaluate their outlier detection performance. Construct $\mathcal{D}_{predictor}={(arch_i, z_i)}_{i=1}^N$ where $arch_i$ is the encoding of the architecture, and $z_i$ is AUROC.
\item Train a neural predictor on $\mathcal{D}_{predictor}$. Use the trained predictor to explore the search space for top-performing CNN architectures.
\item Retrain and evaluate the optimal CNN architecture. % Visualize and identify common patterns in top-performing architectures.
\end{enumerate}

% \paragraph{Searching with Neural Predictor}

% A neural predictor is trained using a carefully curated high-quality dataset, whose construction details will be elaborated in the next section. We employ a multi-layer perceptron as the predictor backbone and the training loss is computed as the sum of the mean squared error and ranking loss. 
% \tunhouc{Can we say regularized evolution?}
% We utilize a modified version of the evolutionary algorithm (EA)\cite{real2019regularized} in conjunction with our trained predictor to comprehensively explore the vast search space for dense connection. Our EA is equipped with two types of mutation steps: either \textit{adding/removing an edge from a selected DAG} or \textit{re-sampling a new DAG to replace the old one}. To overcome the tendency of EA to converge to a local optimal solution, we incorporate an accepting-rejection mechanism inspired by the Metropolis-Hasting algorithm. This allows us to maintain the best structure while still considering weaker candidates based on the difference between their score and the current best score. The acceptance rate is defined as 
% \begin{align}
%     ac = min(1, exp((score^*-score)/T))
% \end{align}
% where $score^*$ and $score$ denote the score of the current best and candidate structure respectively, as evaluated by our predictor.

\subsection{OOD Detection Evaluation}
\begin{figure}[t]
\begin{center}
   \includegraphics[width=\linewidth]{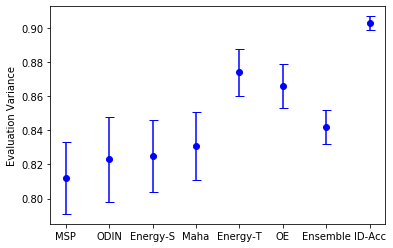}
\end{center}
\vspace{-1em}
\caption{Evaluation variance of sampled networks from search space. Energy(S) and Energy(T) stands for energy scoring and training respectively.}
\label{fig:eval_var}
% \label{fig:onecol}
\end{figure}

In this section, we identify the main challenges in OOD evaluation, and propose the solution to mitigate this issue.

\noindent \textbf{Noisy Evaluation Issue in OOD Detection.}
When using predictor-based NAS, it is crucial to assess the accurate performance of each architecture to build a high-quality dataset for training the predictor. However, for OOD detection performance, the evaluation of neural architectures can be severely noisy. This can lead to inaccurate estimates of the true OOD performance of an architecture and bias the NAS search towards sub-optimal solutions.

 We evaluate the noisy level of several post-training OOD detection methods - MSP, Mahalanobis distance (Maha), ODIN, energy score, and outlier exposure methods - OE, energy training.  To do this, we use CIFAR-10 as our in-distribution dataset and CIFAR-100 as our out-distribution dataset. We randomly sample 10 architectures from our search space and initialize each one with 5 different seeds. We then train the models for 20 epochs by minimizing the cross-entropy loss on the in-distribution data only. For outlier exposure methods, we further fine-tune 5 epochs by using 80M tiny images as auxiliary outlier dataset. We assess the noise level of the trained models by calculating the variance of AUROC on CIFAR-10 versus CIFAR-100, while using in-distribution accuracy on CIFAR-10 as the baseline.

Based on our experiments, we find that the noise level of AUROC is significantly higher than that of ID-accuracy for both post-training and outlier exposure methods. This observation is consistent with our additional experiments, revealing poor Kendall's $\tau$ of the trained predictor for OOD detection. These results suggest that vanilla cross-entropy training with post-training OOD detection methods is strongly impacted by the noisy behavior resulting from the network's random initialization. However, we observe that a few epochs of fine-tuning with outlier exposure methods could substantially reduce the evaluation variance. To build a high-quality predictor, it is crucial to understand the root cause of the significant variance in OOD detection performance. This understanding will enable us to develop a more stable training procedure that can effectively reduce the evaluation noise for NAS.

\paragraph{Multi-view Learning with Outlier Distillation}
% add ensemble variance results?
From a multi-view feature learning perspective, we can gain insight into the behavior of noisy OOD detection evaluation. Vanilla cross-entropy training with different initialization learns a subset of useful features for the classification task. Although different feature subsets may perform similarly on the classification task, they can have varied capabilities in identifying out-of-distribution data, leading to a high variance in OOD detection performance. To mitigate this problem, the outlier exposure method explicitly encourages the model to learn features that can distinguish between in-distribution and OOD samples. While the vanilla OE training aims to produce a uniform distribution on outlier sample, assigning equal probabilities to each class for outlier data can be misleading, particularly for near-OOD samples that are semantically close to the in-distribution data. This can hinder the model's ability to perform effective multi-view feature learning and may also hurt the performance of in-distribution accuracy.

We use Knowledge Distillation~\cite{hinton2015distilling} to provide more accurate and informative supervision from both in-distribution and out-of-distribution training data. Our key idea is to encourage the model to imitate the behavior of a highly performing OOD detection model, enabling it to acquire beneficial multi-view features. Specifically, we have access to a labeled in-distribution training dataset $\Dcal^{train}_{in}$ and an unlabeled outlier dataset $\Dcal^{train}_{out}$. Assume that the neural network use a softmax layer to produces class probabilities on each class $i$:
\begin{align}
    P_i(x) = \frac{\mathrm{exp}(z_i/T)}{\sum_j \mathrm{exp}(z_j/T)},
\end{align}
where $T$ is the temperature scaling parameter to affect the final output distribution. Let $P^t(x)=(P^t_1(x),...,P^t_C(x))$ ($P^s(x)$) denotes the probability output over $C$ classes from the teacher model (student model). The distillation loss is the distance between the teacher and student predictions, given by:

\begin{align}
L^{\mathrm{DS}}(x)=-\sum_{i}[P^t_i(x)\cdot \log P^s_i(x)].
\end{align}
It should be noted that the distillation loss is added to both in-distribution and outlier training data, encouraging the model to imitate the teacher's predictive behavior on both data types. In addition, we also include the cross-entropy loss $L^{\mathrm{CE}}(y;P^s(x))$ for in-distribution data and $L^{\mathrm{CE}}(\Ucal;P^s(x))$ for outlier data respectively, where $\Ucal$ denotes the uniform distribution over $C$ classes. We introduce hyper-parameters $\alpha$ and $\beta$ to control the trade-off between the distillation loss and cross-entropy loss, and use $\lambda$ to balance the contribution of the in-distribution and outlier losses. The final training loss can be expressed as follows:
\begin{align}
L & =\E_{(x,y)\sim \Dcal^{train}_{in}}[\alpha L^{\mathrm{DS}}(x) + (1-\alpha)L^{\mathrm{CE}}(y;P^s(x))] \nonumber \\
&+ \lambda \; \E_{x\sim \Dcal^{train}_{out}}[\beta L^{\mathrm{DS}}(x) + (1-\beta)L^{\mathrm{CE}}(\Ucal;P^s(x))]
 \label{eq_trainL}
\end{align}

\noindent \textbf{Evolving Distillation Algorithm.}
Obtaining a high-performing OOD detection teacher model is crucial in providing valuable knowledge for student learning. Rather than incurring a significant cost to train another model, we can obtain our teacher model by selecting from the sampled networks and using an ensemble of them, see algorithm ~\ref{alg1}. 
% Previous work has demonstrated that model ensembling is highly effective in enhancing OOD performance. 
Specifically, we first sample $N$ architectures from our search space and train them using the vanilla outlier exposure loss. Next, we evaluate the OOD detection performance of the trained architectures on the validation set. Then in each iteration, we select the top-performing architectures and ensemble their outputs as the teacher's output. In practice, we find that a combination of top-performing and randomly selected models can regularize the training and lead to better exploration. We then retrain all architectures using equation ~\ref{eq_trainL} and re-evaluate their performance to prepare for the next iteration. This procedure can be repeated multiple times, resulting in an evolving ensemble of teacher models that provides a progressively improved training loss landscape. At the end of the process, we save $N$ pairs of (arch, AUROC) for predictor training. The final teacher models from last iteration are also saved to provide supervision for re-training. 
% The complete evolving OOD distillation algorithm is summarized in 

\begin{algorithm}[h]
    \caption{Evolving OOD Distillation}
    Sample $N$ architecture, train on $\Dcal^{train}_{in}$ and $\Dcal^{train}_{out}$ with cross-entropy and outlier exposure.
    
    Evaluate trained archs on $\Dcal^{val}$, then add $N$ pair of (arch, AUROC) to $\Hcal$.
    
    \Repeat {maximum iteration is reached} {
        Create teacher set $\Tcal$ by a combination of top $K\%$ and randomly selected $k\%$ archs from $\Hcal$.
        
        Train all $N$ archs with ensemble teacher using equation ~\ref{eq_trainL}
        
        Evaluate trained archs on $\Dcal^{val}$, then update $\Hcal$.
    }
    Use $\Hcal$ for predictor training; Save teacher set $\Tcal$ for later use.
    \label{alg1}
\end{algorithm}

% \begin{algorithm}[H]
%     \caption{Evolving OOD Evaluation}
%     Sample $P$ archs, train on $\Dcal^{train}_{in}$ and $\Dcal^{train}_{out}$ with cross-entropy and outlier exposure.
    
%     Evaluate trained archs on $\Dcal^{val}$, then add $N$ pair of (arch, AUROC) to $\Pcal$ and $\Hcal$.
    
%     \Repeat {maximum iteration is reached} {
%         Create teacher set $\Tcal$ by a combination of top $K\%$ and randomly selected $k\%$ archs from $\Pcal$. 
        
%         Utilize $\Tcal$ to create soft labels (hidden features) on $\Dcal^{train}_{in}$ and $\Dcal^{train}_{out}$, then ensemble with AdaBoost.

%         Sample $q$ new archs, train with ensemble soft-labels (hidden features), evaluate on $\Dcal^{val}$.

%         Update $\Pcal$ by adding $q$ newly evaluated archs and removing 'dead' archs; also add $q$ new archs to $\Hcal$.
        
%     }
%     Use $\Hcal$ for predictor training.
%     \label{alg2}
% \end{algorithm}
% %-------------------------------------------------------------------------

% \begin{figure}[t]
% \begin{center}
% % \fbox{\rule{0pt}{2in} \rule{0.9\linewidth}{0pt}}
%    \includegraphics[width=0.8\linewidth]{Figs/loss_landscape.png}
% \end{center}
%    % \caption{Example of a caption.  It is set in Roman so mathematics
%    % (always set in Roman: $B \sin A = A \sin B$) may be included without an
%    % ugly clash.}
% % \label{fig:long}
% % \label{fig:onecol}
% \end{figure}

\begin{table*}[!ht]
\begin{center}
\resizebox{0.9\textwidth}{!}{
\begin{tabular}{cc|cccc|cccc}
\toprule
 \multicolumn{2}{c}{\multirow{2}{*}{\bf Methods}}
    & \multicolumn{4}{|c|}{\bf CIFAR-10 vs CIFAR-100}  & \multicolumn{4}{c}{\bf CIFAR-100 vs CIFAR-10} \\
    \cline{3-10} 
     & & MSP & Maha & OT & ID-Acc & MSP & Maha & OT & ID-Acc \\ \midrule
\multirow{3}{*}{\bf Standard Arch}
& ResNet-110 & 86.4 & 86.3 &  92.1 & 94.2 & 74.1 & 73.8 & 76.2 & 77.4 \\
& WideResNet-40-2 & 87.5 & 87.7 & 93.0 & \bf95.2 & 74.4 & 74.9 & 75.6 & 78.1  \\
& DenseNet-100-12 & 87.9 & 88.1 & 92.7 & 94.6 & 75.2 & 75.5 & 76.8 & 77.1 \\
\midrule
\multirow{2}{*}{\bf NAS BL}
& Random Search & 88.3 & 88.4 & 92.3 & 94.7 & 76.1 & 76.4 & 78.3 & 77.3 \\
& Regularize EA & 88.9 & 89.2 & 93.2 & 94.9 & 77.1 & 77.5  & 79.5 & 77.9 \\
\midrule
\multirow{2}{*}{\bf Predictor NAS}
& Cross Entropy & 88.1 & 88.3 & 92.4 & 94.8 & 76.2 & 76.3 & 78.1 & 77.5 \\
& DCSOD (Ours) & \bf 90.6 & \bf 91.0 & \bf 95.8(+0.8) & 95.1 & \bf 89.2 & \bf 89.9 & \bf 95.1(+1.1) & \bf 78.3 \\
\bottomrule
\end{tabular}
}    
\end{center}
%\vspace{-2mm}
\caption{Results of OOD AUROC and in-distribution classification accuracy on CIFAR benchmark. OT stands for Outlier Training, i.e. the setting where auxiliary outlier data is used. (.) indicates the improvement over ERD.}
\label{tab:cifar_benchmark}
\end{table*}

\section{Experiments}

In this section, we empirically evaluate the performance of DCSOD on near-OOD detection task. First, we present the detailed configuration of architecture search, evaluation metrics and baselines for comparison. Next, we demonstrate the empirical results on CIFAR benchmarks, followed by a comprehensive ablation study to examine the effect of hyper-parameters. Finally, we visualize the loss landscape and optimal searched cell.

\subsection{Experiment Setup}
\noindent \textbf{Dataset.}
We evaluate our proposed paradigm on both the CIFAR-10 versus CIFAR-100 and CIFAR-100 versus CIFAR-10 detection tasks, where the former is an in-distribution dataset and the latter is an out-of-distribution dataset. We consider two scenarios: with or without auxiliary outlier training data. ("OT" stands for outlier training, corresponding to the scenario with auxiliary outlier data.)  We utilize a modified version of 80 Million Tiny Images as auxiliary outlier training data, wherein all examples that appear in the CIFAR dataset are removed.

\noindent \textbf{Baselines.}
To focus our study on the impact of network architecture, we consider a few popular out-of-distribution (OOD) detection techniques. Specifically, for the case without auxiliary outlier data, we consider the MSP and Mahalanobis distance methods, while for the available auxiliary outlier cases, we consider the vanilla outlier exposure method. Our baseline models consist of three types: (1) widely used human-designed architectures, including ResNet, Wide-ResNet, and DenseNet. (2) NAS baselines include random search and predictor-free regularized EA. (3) predictor-based NAS training using cross-entropy loss. We also compared our results with the current CNN state-of-the-art method ERD~\cite{tifrea2022semi}.

\noindent \textbf{Search Configuration.}
To construct the neural predictor, we randomly sample 2000 architectures from our search space. Each candidate architecture was first pre-trained on 50\% of the CIFAR dataset for 20 epochs, using SGD with an initial learning rate of 0.1 and batch size of 128 via a cosine learning rate schedule. The trained models were evaluated using the aforementioned OOD detection methods. To train for distillation, we select the top 10 performing and 1 randomly selected architectures as teachers. Following the advice from~\cite{hinton2015distilling}, we utilized a high temperature of $T=20$ to facilitate more knowledge transfer from the teacher to the student model. The hyper-parameters chosen for the cases with (without) auxiliary outlier data are $(\lambda,\alpha,\beta)=0.5,0.5,0.2$ ($(\lambda,\alpha)=0, 0.5$), which were determined using validation data. We repeat the iterative process for three times.

The collected dataset, denoted as $\mathcal{D}_{predictor}={(arch_i, auroc_i)}_{i=1}^{2000}$, is split into training data (85\%) and testing data (15\%). We use a multi-layer perceptron as the predictor backbone, and the training loss was computed as the sum of the mean squared error and ranking loss. We employed a modified version of the evolutionary algorithm (EA)\cite{real2019regularized} in conjunction with our trained predictor to comprehensively explore the vast search space for dense connections. Our EA was equipped with two types of mutation steps: either \textit{adding/removing an edge from a selected DAG} or \textit{re-sampling a new DAG to replace the old one}. Finally, we took the optimal searched architecture and re-trained (using previously saved teacher) on the full CIFAR dataset for 300 epochs and evaluate it on the test set.

\noindent \textbf{Evaluation Metrics.}
We assess both the quality of neural predictors and the out-of-distribution (OOD) detection performance of each architecture. To measure the ranking quality provided by the predictor, we use Kendall's Tau and Pearson's Rho. For evaluating OOD detection performance, we use the area under the receiver operating characteristic curve (AUROC), as it exhibits less variance compared to other metrics such as FPR95. Additionally, we report the in-distribution classification accuracy to ensure that the model can maintain its performance on the original task.

\subsection{CIFAR Benchmark Results}
As shown in Table~\ref{tab:cifar_benchmark}, DCSOD consistently outperforms all baselines across various OOD detection methods and configurations. Specifically, our method has improved the previous state-of-the-art (SOTA) ERD on CIFAR by $\sim$1.0\%. For NAS baseline, without auxiliary outlier data, DCSOD improves Regularized EA performance by an average of 2.6\% (13.3\%) for CIFAR-10 vs. CIFAR-100 (CIFAR-100 vs. CIFAR-10). When trained with auxiliary outlier data, the optimal architecture searched by DCSOD outperforms regularized EA by 3.4\% for CIFAR-10 vs. CIFAR-100 (from 92.4\% to 95.8\%), and by 15.6\% for CIFAR-100 vs. CIFAR-10 (from 79.5\% to 95.1\%). However, NAS predictor with cross-entropy training only achieves similar performance as random NAS search due to its low Kendall's Tau, as shown in Table~\ref{tab:predictors}. These results suggest that the neural predictor obtained from DCSOD can more efficiently explore the vast search space for optimal architectures. Additionally, the distillation training loss, constructed by teachers from the evolving distillation algorithm, enhances OOD detection performance and helps maintain classification accuracy.

\begin{table}[!ht]
\centering
\resizebox{0.4\textwidth}{!}{
\begin{tabular}{ccc}
\toprule
 \bf Predictors & \bf Kendall’s $\tau$ & \bf Pearson’s $\rho$\\ \midrule
In-dist Acc & 0.77 & 0.86\\
Cross Entropy-MSP & 0.44 & 0.57 \\
Cross Entropy-OT & 0.51 & 0.62 \\
DCSOD-MSP & 0.72 & 0.81\\
DCSOD-OT & 0.74 & 0.83\\

\bottomrule
\end{tabular}
}
\caption{Evaluation of neural predictor on CIFAR-10 vs CIFAR-100.}
\label{tab:predictors}
\end{table}

\subsection{Ablation Study}
\begin{figure}[t!]%\vspace{-25pt}
    \vspace{-0mm}\centering
    \begin{tabular}{c}
        \includegraphics[height=5cm]{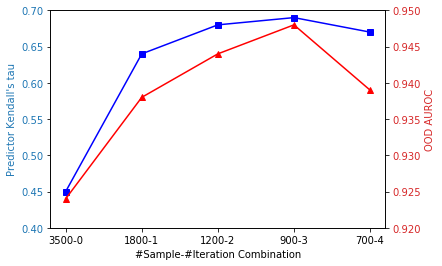}
        % \includegraphics[height=3cm]{Figs/increase_itr.png} \\
        % \hspace{-8mm}
        % \includegraphics[height=3.2cm]{Figs/ce_landscape.pdf} &
        % \hspace{-3mm}
        % \includegraphics[height=3.2cm]{Figs/distill_landscape.pdf} \\
    \end{tabular}
    % \vspace{-2mm}
    \caption{Comparison of out-of-distribution (OOD) AUROC and predictor Kendall's $\tau$ for different sample-iteration trade-offs.}
    % \vspace{-4mm}
    \label{fig:abl1}
\end{figure}

\noindent \textbf{Number of iteration.}
The trade-off between the number of iterations and sample size in the evolving distillation algorithm plays a central role in determining the quality of the neural predictor. To find the appropriate trade-off under a given computational budget, we experimented with multiple iteration-sample size combinations with similar computational costs. For instance, 3000-0 indicates no iteration for teacher selection and distillation training. We conducted experiments on the CIFAR-10 vs. CIFAR-100 task with access to auxiliary outlier data. The results are shown in Figure~\ref{fig:abl1}: vanilla outlier exposure training suffers from a high evaluation variance issue, and the poor quality predictor cannot effectively guide architecture search. With only one iteration of evolving distillation, there is a sharp improvement in both predictor Kendall's $\tau$ and final AUROC performance. From the 1800-1 to 900-3 combinations, a fair improvement can still be witnessed on AUROC, even though Kendall's $\tau$ became saturated. We hypothesize that this is because a better distillation loss term can enhance the final performance of OOD detection. A better version of teacher also reduces the need for a large sample size to train the predictor. However, if we keep reducing the sample size, the predictor will become under-trained without seeing enough examples. We also considered another setup by fixing the sample size to 2000 and increasing the number of iterations. The results show that increasing iterations can consistently increase predictor quality and OOD detection performance, and the improvement starts to saturate after 3 iterations.

\noindent \textbf{Hyper-parameters in Outlier Distillation.}
We conducted a study on the effect of hyper-parameters $\lambda$, $\alpha$, and $\beta$ for outlier distillation training, see Table \ref{tab:hpo}. We began with vanilla cross-entropy training without any distillation loss term. As expected, this setup suffered from high OOD evaluation issues and gave poor performance. After adding the distillation loss on in-distribution data, we observed a significant increase in Kendall's $\tau$. This suggests that imitating the out-performing teacher's behavior only on in-distribution data can facilitate multi-view feature learning and stabilize the OOD detection performance.

Next, we used auxiliary outlier data, but with only $L^{\mathrm{CE}}(\Ucal;P^s(x))$, and observed a noticeable decrease in in-distribution classification accuracy. Finally, we added the distillation loss on outlier data, and the accuracy increased to the level before. This indicates that the outlier distillation term is beneficial to both OOD detection and the original classification task.

\begin{figure}[t!]%\vspace{-25pt}
    \vspace{-0mm}\centering
    \begin{tabular}{cc}
        \hspace{-8mm}
        \includegraphics[height=3.2cm]{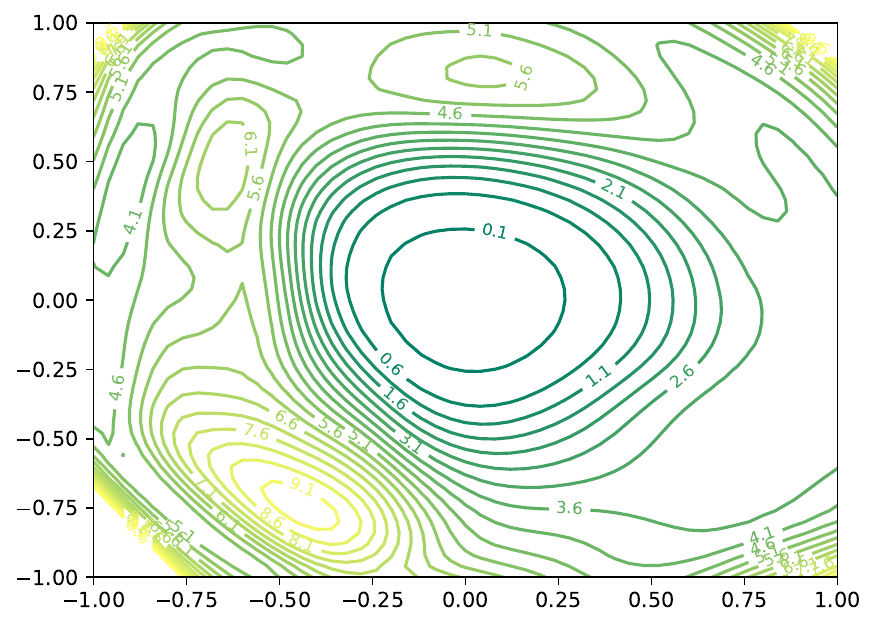} &
        \hspace{-3mm}
        \includegraphics[height=3.2cm]{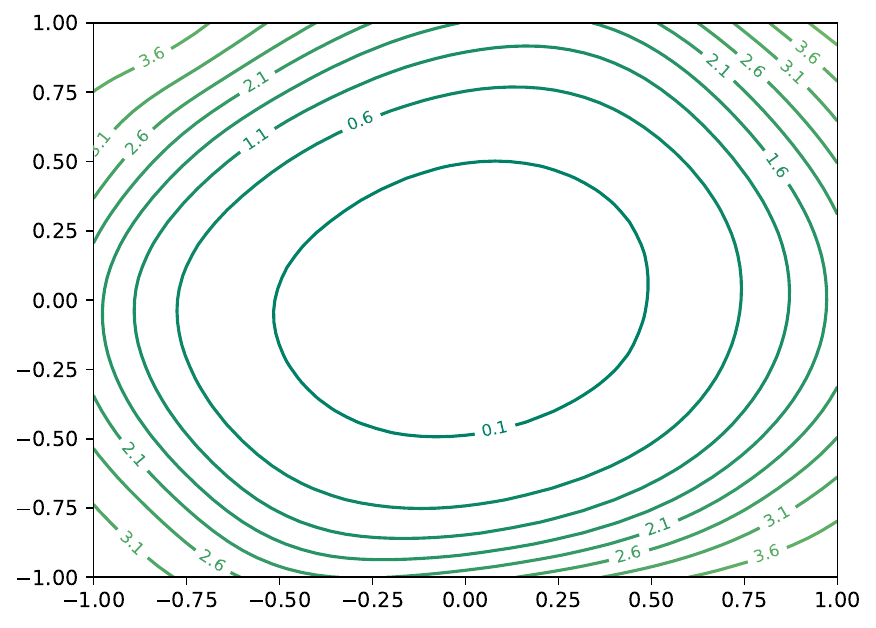} \\
        % \hspace{-8mm}
        % \includegraphics[height=3.2cm]{Figs/ce_landscape.pdf} &
        % \hspace{-3mm}
        % \includegraphics[height=3.2cm]{Figs/distill_landscape.pdf} \\
    \end{tabular}
    % \vspace{-2mm}
    \caption{Visualization of the loss landscape for cross-entropy and outlier distillation loss.}
    % \vspace{-4mm}
    \label{fig:loss_landscape}
\end{figure}

\begin{table}[!ht]
\centering
\resizebox{0.5\textwidth}{!}{
\begin{tabular}{cccc}
\toprule
 \bf {Hyper-parameters} $(\lambda, \alpha, \beta)$ & \bf Kendall’s $\tau$  & \bf AUROC &\bf ID-Acc \\ \midrule
 (0, 0, -) & 0.44 & 88.1 & 94.8 \\
(0, 0.5, -) &  0.67 &  89.4 & 95.1\\
 (0.5, 0.5, 0) & 0.71 &  93.7 &  94.3 \\
 (0.5, 0.5, 0.2) & 0.73 & 95.8 & 95.1 \\
\bottomrule
\end{tabular}
}
\caption{The effect of hyper-parameter $\lambda, \alpha, \beta$}
\label{tab:hpo}
\end{table}

\begin{figure*}[t]%\vspace{-25pt}
    \vspace{-0mm}\centering
    \begin{tabular}{ccc}
        % \hspace{-8mm}
        \includegraphics[height=4.5cm]{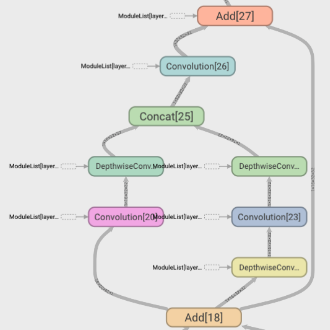} &
        % \hspace{-3mm}
        \includegraphics[height=4.5cm]{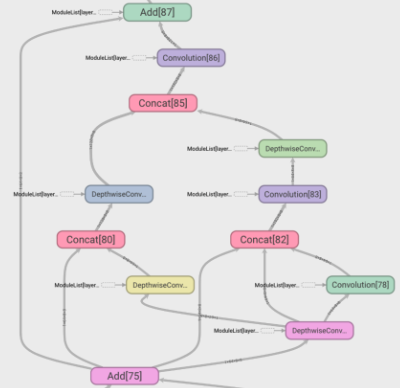} &
        \includegraphics[height=4.5cm]{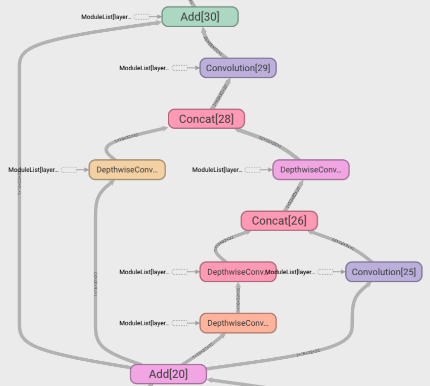}
    \end{tabular}
    % \vspace{-2mm}
    \caption{Visualization of representative cell from DCSOD architectures.}
    % \vspace{-4mm}
    \label{fig:dag_visualize}
\end{figure*}

\subsection{Visualization}

We utilized the "filter normalization" technique \cite{li2018visualizing} to visualize the landscape of the training loss function searched by the evolving distillation algorithm. We took the architecture searched by DCSOD and compared the loss landscape for the vanilla cross-entropy and outlier distillation loss. Specifically, we plotted the loss near each minimizer separately using two random filter-normalized directions. The outlier distillation loss had smoother and wider minima with a convex-like contour and fewer local minima. This suggests that the outlier distillation loss is easier to train and leads to stable evaluation performance on out-of-distribution (OOD) detection.

Last, we visualize several representative cell from top-performing architectures searched by DCSOD, see Fig~\ref{fig:dag_visualize}. 
The discovered models contain versatile structural wirings that effectively aggregate and communicate information within different parts of the CNN architecture, showing superior accuracy/AUROC over hand-crafted arts and NAS-crafted arts. Thus, we answer the question that, allowing dense connectivity within CNN architectures help outlier detection evaluation, and boost the performance of outlier detectors.

% %-------------------------------------------------------------------------
\section{Conclusion}
In this paper, we pioneer the study on the relationship between CNN architectures and their performance on outlier detection. We propose a novel paradigm, \textbf{D}ense \textbf{C}onnectivity \textbf{S}earch of \textbf{O}utlier \textbf{D}etector, that leverages Neural Architecture Search (NAS) to explore the structural wiring of CNN architectures containing dense connectivity of versatile convolution operators. DCSOD utilizes a neural predictor to guide the search, and improves the evaluation of candidate outlier detectors by proposing evolving distillation algorithm. Evaluation results show that our NAS-crafted models achieve the state-of-the-art performance on CIFAR benchmarks, distinguishing out-of-distribution examples from CIFAR-10/CIFAR-100 benchmarks with higher accuracy and AUROC, answering the question that dense connectivity is beneficial for the construction of outlier detectors.

% %------------------------------------------------------------------------
% \section{Final copy}

% You must include your signed IEEE copyright release form when you submit
% your finished paper. We MUST have this form before your paper can be
% published in the proceedings.

{\small
\bibliographystyle{ieee_fullname}
\bibliography{egbib}
}

\end{document}